\documentclass[runningheads]{llncs}

\usepackage[T1]{fontenc}
\usepackage{graphicx}

\usepackage{cite}
\usepackage{hyperref}
\usepackage[utf8]{inputenc}
\usepackage{framed,multirow}
\usepackage{tikz}
\usetikzlibrary{arrows,shapes,positioning,shadows,trees,calc,decorations.markings}
\usepackage{pifont}
\usepackage{multirow}
\usepackage{booktabs}
\usepackage{amsmath}
\usepackage{amssymb}

\begin{document}

\title{Surgical-VQA: Visual Question Answering in Surgical Scenes using Transformer}
\titlerunning{Surgical-VQA}

\author{Lalithkumar Seenivasan\inst{1, \star} \orcidID{0000-0002-0103-1234} \and
Mobarakol Islam\inst{2, }\thanks{Lalithkumar Seenivasan and Mobarakol Islam are co-first authors.}\orcidID{0000-0002-7162-2822} \and
Adithya K Krishna\inst{3}\orcidID{0000-0002-2284-703X} \and
Hongliang Ren\inst{1, 4, 5, }\thanks{Corresponding author.}\orcidID{0000-0002-6488-1551}}
% index{Seenivasan, Lalithkumar} 
% index{Islam, Mobarakol} 
% index{Krishna, Adithya K}
% index{Ren, Hongliang}

\authorrunning{Seenivasan et al.}

\institute{
Dept. of Biomedical Engineering, National University of Singapore, Singapore. \and
Biomedical Image Analysis Group, Imperial College London, UK. \and
Dept. of ECE, National Institute of Technology, Tiruchirappalli, India. \and
Dept. of Electronic Engineering, Chinese University of Hong Kong. \and
Shun Hing Institute of Advanced Engineering, Chinese University of Hong Kong.\\
\email{lalithkumar\_s@u.nus.edu, m.islam20@imperial.ac.uk, 108118004@nitt.edu, ren@nus.edu.sg/hlren@ee.cuhk.edu.hk}
}

\maketitle              % typeset the header of the contribution

\begin{abstract}

Visual question answering (VQA) in surgery is largely unexplored. Expert surgeons are scarce and are often overloaded with clinical and academic workloads. This overload often limits their time answering questionnaires from patients, medical students or junior residents related to surgical procedures. At times, students and junior residents also refrain from asking too many questions during classes to reduce disruption. While computer-aided simulators and recording of past surgical procedures have been made available for them to observe and improve their skills, they still hugely rely on medical experts to answer their questions. Having a Surgical-VQA system as a reliable ‘second opinion’ could act as a backup and ease the load on the medical experts in answering these questions. The lack of annotated medical data and the presence of domain-specific terms has limited the exploration of VQA for surgical procedures. In this work, we design a Surgical-VQA task that answers questionnaires on surgical procedures based on the surgical scene. Extending the  MICCAI endoscopic vision challenge 2018 dataset and workflow recognition dataset further, we introduce two Surgical-VQA datasets with classification and sentence-based answers. To perform Surgical-VQA, we employ vision-text transformers models. We further introduce a residual MLP-based VisualBert encoder model that enforces interaction between visual and text tokens, improving performance in classification-based answering. Furthermore, we study the influence of the number of input image patches and temporal visual features on the model performance in both classification and sentence-based answering.

% \keywords{Surgical-VQA  \and Visual Question Answering \and Robotic Surgery \and Vision-text encoder.}

\end{abstract}

\section{Introduction} \label{introduction}

Lack of medical domain-specific knowledge has left many patients, medical students and junior residents with questions lingering in their minds about medical diagnosis and surgical procedures. Many of these questions are left unanswered either because they assume these questions to be thoughtless, or students and junior residents refrain from raising too many questions to limit disruptions in lectures. The chances for them finding a medical expert to clarify their doubts are also slim due to the scarce number of medical experts who are often overloaded with clinical and academic works~\cite{bates2000error}. To assist students in sharpening their skills in surgical procedures, many computer-assisted techniques~\cite{adams1990computer, rogers1998computer} and simulators~\cite{kneebone2003simulation, sarker2007simulation} have been proposed. Although the systems assist in improving their skills and help reduce the workloads on academic professionals, the systems don’t attempt to answer the student’s doubts. While students have also been known to learn by watching recorded surgical procedures, the tasks of answering their questions still fall upon the medical experts. In such cases, a computer-assisted system that can process both the medical data and the questionnaires and provide a reliable answer would greatly benefit the students and reduce the medical expert’s workload~\cite{sharma2021medfusenet}. Surgical scenes are enriched with information that the system can exploit to answer questionnaires related to the defective tissue, surgical tool interaction and surgical procedures.

With the potential to extract diverse information from a single visual feature just by varying the question, the computer vision domain has seen a recent influx of vision and natural language processing models for visual question answering (VQA) tasks~\cite{sheng2021human,wang2021simvlm,li2019visualbert}. These models are either built based on the long short-term memory (LSTM)~\cite{sharma2021image, barra2021visual} or attention modules~\cite{sharma2021visual, zhang2021multimodal, sharma2021medfusenet}. In comparison to the computer vision domain, which is often complemented with massive annotated datasets, the medical domain suffers from the lack of annotated data, limiting the exploration of medical VQA. The presence of domain-specific medical terms also limits the use of transfer learning techniques to adapt pre-trained computer-vision VQA models for medical applications. While limited works have been recently reported on medical-VQA~\cite{sharma2021medfusenet} for medical diagnosis, VQA for surgical scenes remains largely unexplored.

\begin{figure}[!b]
    \centering
    \includegraphics[width=1.0\textwidth]{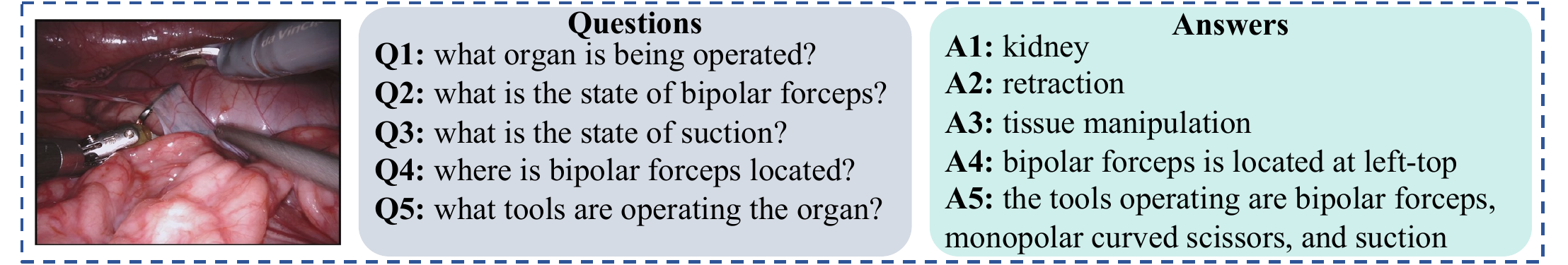}
    \caption{Surgical-VQA: Given a surgical scene, the model predicts answer related to surgical tools, their interactions and surgical procedures based on the questionnaires.}
    \label{fig:surgical_VQA}
\end{figure}

In this work, \textbf{(i)} we design a Surgical-VQA task to generate answers for questions related to surgical tools, their interaction with tissue and surgical procedures (Fig.~\ref{fig:surgical_VQA}). \textbf{(ii)} We exploit the surgical scene segmentation dataset from the MICCAI endoscopic vision challenge 2018 (EndoVis-18)~\cite{allan20202018} and workflow recognition challenge dataset (Cholec80)~\cite{twinanda2016endonet}, and extend it further to introduce two novel datasets for Surgical-VQA tasks. \textbf{(iii)} We employ two vision-text attention-based transformer models to perform classification-based and sentence-based answering for the Surgical-VQA. \textbf{(iv)} We also introduce a residual MLP (ResMLP) based VisualBERT ResMLP encoder model that outperforms VisualBERT~\cite{li2019visualbert} in classification-based VQA. Inspired by ResMLP~\cite{touvron2021resmlp}, cross-token and cross-channel sub-modules are introduced into the VisualBERT ResMLP model to enforce interaction among input visual and text tokens. \textbf{(v)} Finally, the effects on the model’s performance due to the varied number of input image patches and inclusion of temporal visual features are also studied.

\section{Proposed Method} \label{proposed_method}

\subsection{Preliminaries} \label{preliminaries}

\textbf{VisualBERT~\cite{li2019visualbert}:} 
A multi-layer transformer encoder model that integrates BERT~\cite{devlin2018bert} transformer model with object proposal models to perform vision-and-language tasks. BERT~\cite{devlin2018bert} model primarily processes an input sentence as a series of tokens (subwords) for natural language processing. By mapping to a set of embeddings ($E$), each word token is embedded ($e \in E$)  based on token embedding $e_t$, segment embedding $e_s$ and position embedding $e_p$. Along with these input word tokens, VisualBERT~\cite{li2019visualbert} model processes visual inputs as unordered visual tokens that are generated using the visual features extracted from the object proposals. In addition to text embedding from BERT~\cite{devlin2018bert}, it performs visual embedding ($F$), where, each visual token is embedded ($f \in F$) based on visual features $f_o$, segment embedding $f_s$ and position embedding $f_p$. Both text and visual embedding are then propagated through multiple encoder layers in the VisualBERT~\cite{li2019visualbert} to allow rich interactions between both the text and visual tokens and establish joint representation. Each encoder layer consists of an (i) self-attention module that establishes relations between tokens, (ii) intermediate module and (iii) output module consisting of hidden linear layers to reason across channels. Finally, the encoder layer is followed by a pooler module.

\begin{figure}[!b]
    \centering
    \includegraphics[width=1.0\textwidth]{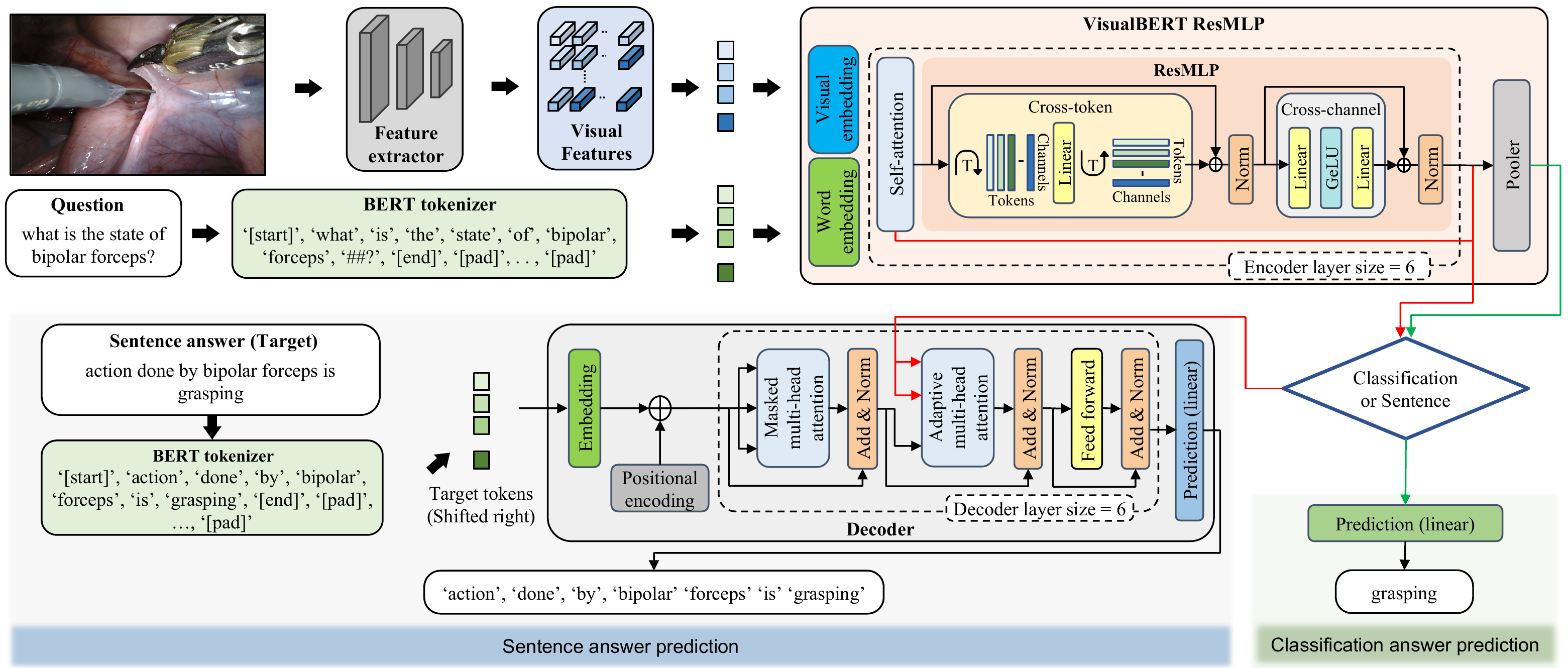}
    \caption{Architecture: Given an input surgical scene and questions, its text and visual features are propagated through the vision-text encoder (VisualBERT ResMLP). \textbf{(i) classification-based answer:} The encoder output is propagated through a prediction layer for answer classification. \textbf{(ii) Sentence-based answer:} The encoder is combined with a transformer decoder to predict the answer sentence word-by-word (regressively).}
    \label{fig:architecture}
\end{figure}

\subsection{VisualBERT ResMLP} \label{VisualBertResMLP}

In our proposed VisualBERT ResMLP encoder model, we aim to further boost the interaction between the input tokens for vision-and-language tasks. The VisualBERT~\cite{li2019visualbert} model relies primarily on its self-attention module in the encoder layers to establish dependency relationships and allow interactions among tokens. Inspired by residual MLP (ResMLP)~\cite{touvron2021resmlp}, the intermediate and output modules of the VisualBERT~\cite{li2019visualbert} model are replaced by cross-token and cross-channel modules to further enforce interaction among tokens. In the cross-token module, the inputs word and visual tokens are transposed and propagated forward, allowing information exchange between tokens. The resultant is then transposed back to allow per-token forward propagation in the cross-channel module. Both cross-token and cross-channel modules are followed by element-wise summation with a skip-connection (residual-connection), which are layer-normalized. The cross-token output ($X_{CT}$) and cross-channel output($X_{CC}$) is theorised as:

% \begin{equation}
%   X_{CT} = Norm( X_{SA} + (A((X_{SA})^{T}))^{T})
% \label{eq:cross_patch}
% \end{equation}

% \begin{equation}
%   X_{CC} = Norm( X_{CT} + C(GeLU(B(X_{CT}))))
% \label{eq:cross_channel}
% \end{equation}

\begin{flalign}
X_{CT} &= Norm( X_{SA} + (A((X_{SA})^{T}))^{T})\\
X_{CC} &= Norm( X_{CT} + C(GeLU(B(X_{CT}))))
\end{flalign}

\noindent where, $X_{SA}$ is the self-attention module output, $A$, $B$ and $C$ are the learnable linear layers, $GeLU$ is the GeLU activation function~\cite{hendrycks2016gaussian} and NORM is the layer-normalization.

\subsection{VisualBert ResMLP for Classification} \label{VB_ResMLP_Cls}

\textbf{Word and Visual Tokens:} 
Each question is converted to a series of word tokens generated using BERT tokenizer~\cite{devlin2018bert}. Here, the BERT tokenizer is custom trained on the dataset to include surgical domain-specific words. The visual tokens are generated using the final convolution layer features of the feature extractor. A ResNet18~\cite{he2016deep} model pre-trained on ImageNet~\cite{deng2009imagenet} is used as the feature extractor. While the VisualBERT~\cite{li2019visualbert} model uses the visual features extracted from object proposals, we bypass the need for object proposal networks by extracting the features from the entire image. By employing adaptive average pooling to the final convolution layer, the output shape (s) is resize to s = [batch size x $n$ x $n$ x $256$], thereby, restricting the number of visual tokens (patches) to $n^2$. 
\textbf{VisualBERT ResMLP Module:}
The word and visual tokens are propagated through text and visual embedding layers, respectively, in the VisualBert ResMLP model. The embedded tokens are then propagated through 6 layers of encoders (comprising self-attention and ResMLP modules) and finally through the pooling module. Embedding size = 300 and hidden layer feature size = 2048 are set as VisualBert ResMLP model parameters.
\textbf{Classification-based answer:}
The output of the pooling module is propagated through a prediction (linear) layer to predict the answer label.

\subsection{VisualBert ResMLP for Sentence} \label{VB_ResMLP_Sen}

\textbf{Word and Visual Tokens:} In addition to tokenizing the words in the questions and visual features in the image as stated in section~\ref{VB_ResMLP_Cls}, the target sentence-based answer is also converted to a series of word tokens using the BERT tokenizer.
\textbf{VisualBERT ResMLP encoder:}
The VisualBERT ResMLP model follows the same parameters and configuration as stated in section~\ref{VB_ResMLP_Cls}.
\textbf{Sentence-based answer:} To generate the answer in a sentence structure, we propose to combine the vision-text encoder model (VisualBERT~\cite{li2019visualbert} or VisualBERT ResMLP) with a multi-head attention-based transformer decoder (TD) model. Considering the sentence-based answer generation task to be similar to an image-caption generation task~\cite{cornia2020meshed}, it is positioned as a next-word prediction task. Given the question features, visual features and the previous word (in answer sentence) token, the decoder model predicts the next word in the answer sentence. During the training stage, as shown in Fig.~\ref{fig:architecture}, the target word tokens are shifted rights and, together with vision-text encoder output, are propagated through the transformer decoder layers to predict the corresponding next word. During the evaluation stage, the beam search~\cite{wiseman2016sequence} technique is employed to predict the sentence-based answer without using target word tokens. Taking the ‘[start]’ token as the first word, the visual-text encoder model and decoder is regressed on every predicted word to predict the subsequent word, until a ‘[end]’ token is predicted.

\section{Experiment} \label{experiments}

\subsection{Dataset} \label{dataset}

\subsubsection{Med-VQA:}

A public dataset from the ImageCLEF 2019 Med-VQA Challenge~\cite{abachavqa}. Three categories (C1: modality, C2: plane and C3: organ) of medical question-answer pairs from the dataset are used in this work. The C1, C2 and C3 pose a classification task (single-word answer) for 3825 images with a question. The C1, C2 and C3 consist of 45, 16 and 10 answer classes, respectively. The train and test set split follows the original implementation~\cite{abachavqa}.

\subsubsection{EndoVis-18-VQA:}

A novel dataset was generated from $14$ video sequences of robotic nephrectomy procedures from the MICCAI Endoscopic Vision Challenge 2018~\cite{allan20202018} dataset. Based on the tool, tissue, bounding box and interaction annotations used for tool-tissue interaction detection tasks~\cite{islam2020learning}, we generated two versions of question-answer pairs for each image frame. \textbf{(i) Classification (EndoVis-18-VQA (C)):} The answers are in single-word form. It consists of $26$ distinct answer classes (one organ, $8$ surgical tools, $13$ tool interactions and $4$ tool locations). \textbf{(ii) Sentence (EndoVis-18-VQA (S)):} The answers are in a sentence form. In both versions, $11$ sequences with $1560$ images and $9014$ question-answer pairs are used as a training set and $3$ sequences with $447$ images and $2769$ question-answers pairs are used as a test set. The train and test split follow the tool-tissue interaction detection task~\cite{seenivasan2022global}.

\subsubsection{Cholec80-VQA:}

A novel dataset generated from $40$ video sequences of the Cholec80 dataset~\cite{twinanda2016endonet}. We sampled the video sequences at $0.25$ fps to generate the Cholec80-VQA dataset consisting of 21591 frames. Based on tool-operation and phase annotations provided in the Cholec80 dataset~\cite{twinanda2016endonet}, two question-answer pairs are generated for each frame. \textbf{(i) Classification (Cholec80-VQA (C)):} $14$ distinct single-word answers (8 on surgical phase, two on tool state and 4 on number of Tools). \textbf{(ii) Sentence (Cholec80-VQA (S)):} The answers are in a sentence form. In both versions, $34$k question-answer pairs for $17$k frames are used for the train set and $9K$ question-answer pairs for $4.5$k frames are used for the test set. The train and test set split follows the Cholec80 dataset~\cite{twinanda2016endonet}.

\subsection{Implementation Details}

Both our classification and the sentence-based answer generation models\footnote[1]{\href{https://github.com/lalithjets/Surgical_VQA.git}{github.com/lalithjets/Surgical\_VQA.git}} are trained based on cross-entropy loss and optimized using the Adam optimizer. For Classification tasks, a batch size = 64 and epoch = 80 are used. A learning rate  = $5$x$10^{-6}$, $1$x$10^{-5}$ and $5$x$ 10^{-6}$ are used for  Med-VQA\cite{abachavqa}, EndoVis-18-VQA (C) and Cholec80-VQA (C) dataset, respectively. For the sentence-based answer generation tasks, a batch size = 50 is used. The models are trained for epoch = 50, 100 and 51, with a learning rate  = $1$x$10^{-4}$, $5$x$10^{-5}$ and $1$x$ 10^{-6}$ on Med-VQA\cite{abachavqa}, EndoVis-18-VQA (S) and Cholec80-VQA (S) dataset, respectively.

\section{Results}

The performance of classification-based answering on the EndoVis-18-VQA  (C), Cholec80-VQA (C) and Med-VQA (C1, C2 and C3) datasets are quantified based on the accuracy (Acc), recall and F-score in Table~\ref{table:classification}. It is observed that our proposed encoder (VisualBERT ResMLP) based model outperformed the current medical VQA state-of-the-art MedFuse~\cite{sharma2021medfusenet} model and marginally outperformed the base encoder (VisualBERT~\cite{li2019visualbert}) based model in almost all datasets. While the improvement in performance against the base model is marginal, a k-fold study (Table~\ref{table:kfold_test}) on EndoVis-18-VQA  (C) dataset proves that the improvement is consistent. Furthermore, our model ($159.0$M) requires $13.64$\% lesser parameters compared to the base model ($184.2$M).

\begin{table*}[!t]
    \centering
    \caption{Performance comparison of our VisualBERT ResMLP based model against MedFuse~\cite{sharma2021medfusenet} and VisualBERT~\cite{li2019visualbert} based model for classification-based answering.}
    \scalebox{0.75}{
    \begin{tabular}{c|ccccccccc}
        \toprule
        \multirow{2}{*}{\textbf{Dataset}}   & \multicolumn{3}{c|}{\textbf{MedFuse~\cite{sharma2021medfusenet}}}                                               & \multicolumn{3}{c|}{\textbf{VisualBert~\cite{li2019visualbert}}}                                                         & \multicolumn{3}{c}{\textbf{VisualBert ResMLP}}                                                      \\ \cline{2-10} 
                                            & \multicolumn{1}{c|}{\textbf{Acc}} & \multicolumn{1}{c|}{\textbf{Recall}} & \multicolumn{1}{c|}{\textbf{Fscore}} & \multicolumn{1}{c|}{\textbf{Acc}}      & \multicolumn{1}{c|}{\textbf{Recall}}   & \multicolumn{1}{c|}{\textbf{Fscore}}   & \multicolumn{1}{c|}{\textbf{Acc}}      & \multicolumn{1}{c|}{\textbf{Recall}}   & \textbf{Fscore}   \\
        \midrule
        \textbf{Med-VQA (C1)}               & \multicolumn{1}{c}{0.754}         & \multicolumn{1}{c}{0.224}            & \multicolumn{1}{c|}{0.140}           & \multicolumn{1}{c}{\textbf{0.828}}     & \multicolumn{1}{c}{\textbf{0.617}}     & \multicolumn{1}{c|}{\textbf{0.582}}    & \multicolumn{1}{c}{\textbf{0.828}}     & \multicolumn{1}{c}{0.598}              & 0.543             \\ %\hline
        \textbf{Med-VQA (C2)}               & \multicolumn{1}{c}{0.730}         & \multicolumn{1}{c}{0.305}            & \multicolumn{1}{c|}{0.303}           & \multicolumn{1}{c}{\textbf{0.760}}     & \multicolumn{1}{c}{0.363}              & \multicolumn{1}{c|}{0.367}             & \multicolumn{1}{c}{0.758}              & \multicolumn{1}{c}{\textbf{0.399}}     & \textbf{0.398}    \\ %\hline
        \textbf{Med-VQA (C3)}               & \multicolumn{1}{c}{0.652}         & \multicolumn{1}{c}{0.478}            & \multicolumn{1}{c|}{0.484}           & \multicolumn{1}{c}{0.734}              & \multicolumn{1}{c}{0.587}              & \multicolumn{1}{c|}{0.595}             & \multicolumn{1}{c}{\textbf{0.736}}     & \multicolumn{1}{c}{\textbf{0.609}}     & \textbf{0.607}    \\ %\hline
        \textbf{EndoVis-18-VQA (C)}         & \multicolumn{1}{c}{0.609}         & \multicolumn{1}{c}{0.261}            & \multicolumn{1}{c|}{0.222}           & \multicolumn{1}{c}{0.619}              & \multicolumn{1}{c}{\textbf{0.412}}     & \multicolumn{1}{c|}{0.334}             & \multicolumn{1}{c}{\textbf{0.632}}     & \multicolumn{1}{c}{0.396}              & \textbf{0.336}    \\ %\hline
        \textbf{Cholec80-VQA (C)}           & \multicolumn{1}{c}{0.861}         & \multicolumn{1}{c}{0.349}            & \multicolumn{1}{c|}{0.309}           & \multicolumn{1}{c}{0.897}              & \multicolumn{1}{c}{\textbf{0.629}}     & \multicolumn{1}{c|}{0.633}             & \multicolumn{1}{c}{\textbf{0.898}}     & \multicolumn{1}{c}{0.627}              & \textbf{0.634}    \\
        \bottomrule
    \end{tabular}
    }
    \label{table:classification}
\end{table*}

\begin{table}[!t]
    \centering
    \caption{k-fold performance comparison of our VisualBERT ResMLP based model against VisualBERT~\cite{li2019visualbert} based model on EndoVis-18-VQA (C) dataset.}
    \scalebox{0.75}{
    \begin{tabular}{c|cc|cc|cc}
        \toprule
        \multirow{2}{*}{\textbf{Model}}                 & \multicolumn{2}{c|}{\textbf{$1^{st}$ Fold}}               & \multicolumn{2}{c|}{\textbf{$2^{nd}$ Fold}}               & \multicolumn{2}{c}{\textbf{$3^{rd}$ Fold}}            \\ \cline{2-7} 
                                                        & \multicolumn{1}{c|}{\textbf{Acc}}     & \textbf{Fscore}   & \multicolumn{1}{c|}{\textbf{Acc}}   & \textbf{Fscore}     & \multicolumn{1}{c|}{\textbf{Acc}}   & \textbf{Fscore} \\
        \midrule
        \textbf{VisualBert~\cite{li2019visualbert}}     & \multicolumn{1}{c}{0.619}             & 0.334             & \multicolumn{1}{c}{0.605}           & 0.313               & \multicolumn{1}{c}{0.578}           & 0.337           \\
        \textbf{VisualBert ResMLP}                      & \multicolumn{1}{c}{\textbf{0.632}}    & \textbf{0.336}    & \multicolumn{1}{c}{\textbf{0.649}}  & \textbf{0.347}      & \multicolumn{1}{c}{\textbf{0.585}}  & \textbf{0.373}  \\
        \bottomrule
    \end{tabular}
    }
    \label{table:kfold_test}
\end{table}

\begin{table*}[!b]
    \centering
    \caption{Comparison of transformer-based models ((i) VisualBERT~\cite{li2019visualbert} + TD and (ii) VisualBERT ResMLP +  TD) against MedFuse~\cite{sharma2021medfusenet} for sentence-based answering.}
    \scalebox{0.74}{
        \begin{tabular}{c|cccc|cccc}
            \toprule
            \multirow{2}{*}{\textbf{Model}}                     & \multicolumn{4}{c|}{\textbf{EndoVis-18-VQA (S)}}                                                                                     & \multicolumn{4}{c}{\textbf{Cholec80-VQA (S)}}                                                                                       \\ \cline{2-9} 
                                                                & \multicolumn{1}{c|}{\textbf{BLEU-3}} & \multicolumn{1}{c|}{\textbf{BLEU-4}} & \multicolumn{1}{c|}{\textbf{CIDEr}} & \textbf{METEOR}  & \multicolumn{1}{c|}{\textbf{BLEU-3}} & \multicolumn{1}{c|}{\textbf{BLEU-4}} & \multicolumn{1}{c|}{\textbf{CIDEr}} & \textbf{METEOR} \\
            \midrule
            \textbf{MedFuse~\cite{sharma2021medfusenet}}        & \multicolumn{1}{c}{0.212}            & \multicolumn{1}{c}{0.165}            & \multicolumn{1}{c}{0.752}           & 0.148            & \multicolumn{1}{c}{0.378}            & \multicolumn{1}{c}{0.333}            & \multicolumn{1}{c}{1.250}            & 0.222          \\ %\hline
            \textbf{VisualBert~\cite{li2019visualbert} + TD}    & \multicolumn{1}{c}{\textbf{0.727}}   & \multicolumn{1}{c}{\textbf{0.694}}   & \multicolumn{1}{c}{5.153}           & \textbf{0.544}   & \multicolumn{1}{c}{\textbf{0.963}}   & \multicolumn{1}{c}{\textbf{0.956}}   & \multicolumn{1}{c}{\textbf{8.802}}   & \textbf{0.719} \\ %\hline
            \textbf{VisualBert ResMLP + TD}                     & \multicolumn{1}{c}{0.722}            & \multicolumn{1}{c}{0.691}            & \multicolumn{1}{c}{\textbf{5.262}}  & 0.543            & \multicolumn{1}{c}{0.960}            & \multicolumn{1}{c}{0.952}            & \multicolumn{1}{c}{8.759}            & 0.711          \\
            \bottomrule
        \end{tabular}
    }
    \label{table:sentence}
\end{table*}

\begin{figure}[!b]
    \centering
    \includegraphics[width=1\textwidth]{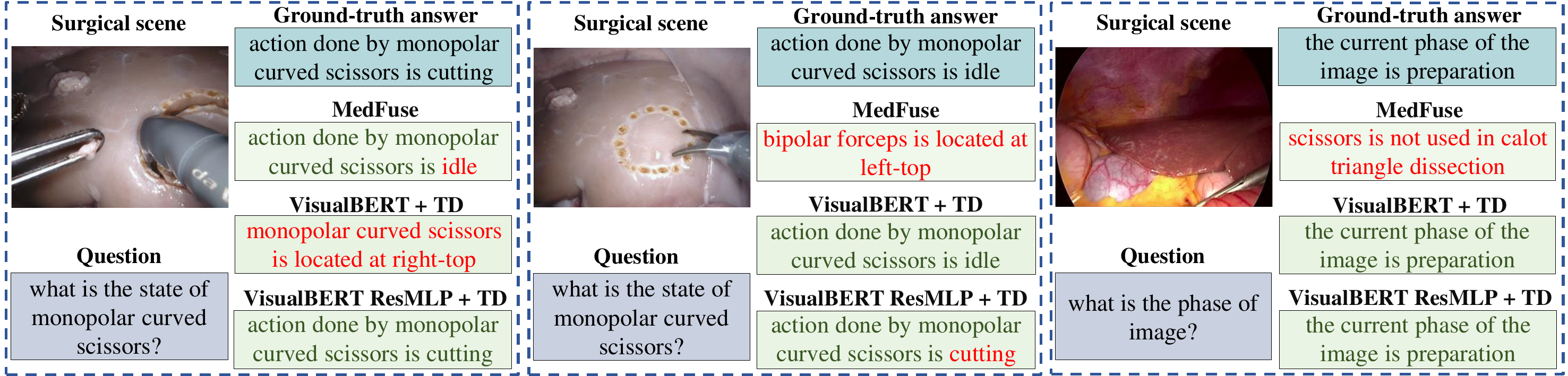}
    \caption{Comparison of sentence-based answer generation by MedFuse~\cite{sharma2021medfusenet}, VisualBERT~\cite{li2019visualbert} + TD and VisualBERT ResMLP + TD models.}
    \label{fig:qualitative_analysis}
\end{figure}

For the sentence-based answering, BLEU score~\cite{papineni2002bleu}, CIDEr~\cite{vedantam2015cider} and METEOR~\cite{banerjee2005meteor} are used for quantitative analysis. Compared to the LSTM-based MedFuse~\cite{sharma2021medfusenet} model, the two variants of our proposed Transformer-based model (VisualBERT~\cite{li2019visualbert} + TD and VisualBERT ResMLP + TD) performed better on both the EndoVis-18-VQA (S) and Cholec80-VQA (S) dataset (Table ~\ref{table:sentence}). However, when compared within the two variants, the variant with the VisualBERT~\cite{li2019visualbert} as its vision-text encoder performed marginally better. While the cross-patch sub-module in VisualBERT ResMLP encoder improves performance in classification-based answering, the marginal low performance could be attributed to its influence on the self-attention sub-module. To predict a sentence-based answer, in addition to the encoder’s overall output, the adaptive multi-head attention sub-module in the TD utilizes the encoder’s self-attention sub-module outputs. By enforcing interaction between tokens, the cross-patch sub-module could have affected the optimal training of the self-attention sub-module, thereby affecting the sentence generation performance. It is worth noting that the VisualBERT ResMLP encoder + TD model ($184.7$M) requires 11.98\% fewer parameters compared to the VisualBERT~\cite{li2019visualbert} + TD model ($209.8$M) while maintaining similar performances. Fig.~\ref{fig:qualitative_analysis} shows the qualitative performance of sentence-based answering.

\subsection{Ablation Study}

\begin{figure}[!b]
    \centering
    \includegraphics[width=1.0\textwidth]{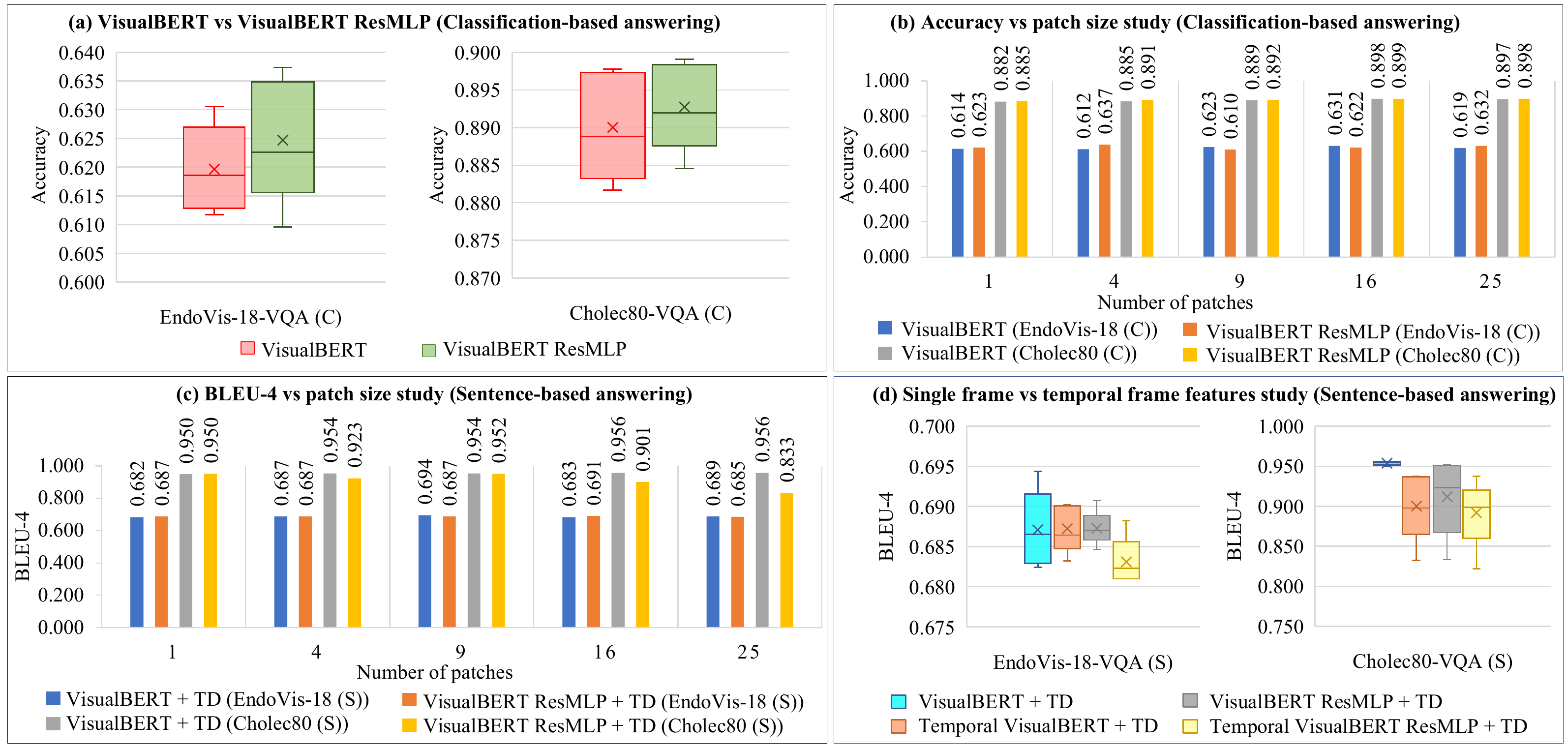}
    \caption{Ablation study: For classification-based answering (a) Performance comparison of VisualBERT~\cite{li2019visualbert}  vs VisualBERT ResMLP based model; (b) Accuracy vs patch size (number of patches) study. For sentence-based answering: (c) BLEU-4 vs patch size study; (d) Performance of single frame vs temporal visual features.}
    \label{fig:ablation_study}
\end{figure}

Firstly, the performance of VisualBERT encoder and VisualBERT ResMLP encoder-based models for classification-based answering with a varying number of input image patches (patch size = 1, 4, 9, 16 and 25) is trained and studied. From Fig.~\ref{fig:ablation_study} (a), it is observed that VisualBERT ResMLP encoder-based model generally performs better (in-line with observation in Table~\ref{table:classification})  than VisualBERT~\cite{li2019visualbert} encoder-based model, even with varied number of input patches. Secondly, from Fig.~\ref{fig:ablation_study} (b), it is also observed that performances generally improved with an increase in the number of patches for classification-based answering. However, the influence of the number of inputs patches on sentence-based answering remains inconclusive (Fig.~\ref{fig:ablation_study} (c)). In most cases, the input of a single patch seems to offer the best or near best results.

Finally, the performance of models incorporated with temporal visual features is also studied. Here, a 3D ResNet50 pre-trained on the UCF101 dataset ~\cite{hara2017learning} is employed as the feature extractor. With a temporal size = 3, the current and the past 2 frames are used as input to the feature extractor. The extracted features are then used as visual tokens for sentence-based answering. Fig.~\ref{fig:ablation_study} (d) shows the model’s performance on a varied number of input patches (patch size = 1, 4, 9, 16 and 25) from single frames vs. temporal frames on EndoVis-18-VQA (S) and Cholec80-VQA (S). It is observed that both transformer-based models' performance reduces when the temporal features are used.

\section{Discussion and Conclusion}

We design a Surgical-VQA algorithm to answer questionnaires on surgical tools, their interactions and surgical procedures based on our two novel Surgical-VQA datasets evolved from two public datasets. To perform classification and sentence-based answering, vision-text attention-based transformer models are employed. A VisualBERT ResMLP transformer encoder model with lesser model parameters is also introduced that marginally outperforms the base vision-text attention encoder model by incorporating a cross-token sub-module. The influence of the number of input image patches and the inclusion of temporal visual features on the model’s performance is also reported. While our Surgical-VQA task answers to less-complex questions, from the application standpoint, it unfolds the possibility of incorporating open-ended questions where the model could be trained to answer surgery-specific complex questionnaires. From the model standpoint, future work could focus on introducing an asynchronous training regime to incorporate the benefits of the cross-patch sub-module without affecting the self-attention sub-module in sentence-based answer-generation tasks.

\section{Acknowledgement}

We thank Ms. Xu Mengya, Dr. Preethiya Seenivasan and Dr. Sampoornam Thamizharasan for their valuable inputs during project discussions. This work is supported by Shun Hing Institute of Advanced Engineering (SHIAE project BME-p1-21, 8115064) at the Chinese University of Hong Kong (CUHK), Hong Kong Research Grants Council (RGC) Collaborative Research Fund (CRF C4026-21GF and CRF C4063-18G) and (GRS)\#3110167.

\bibliographystyle{splncs04}
\bibliography{paper0916}

\end{document}